\documentclass[conference]{IEEEtran}
\IEEEoverridecommandlockouts
\usepackage{cite}
\usepackage{amsmath,amssymb,amsfonts}
\usepackage{graphicx}
\usepackage{textcomp}
\usepackage{xcolor}
\usepackage{algorithm}
\usepackage{algpseudocode}
\usepackage{multirow}
\usepackage{stfloats}
\usepackage{booktabs}
\usepackage{siunitx}
\usepackage{tabularx}
\usepackage[utf8]{inputenc}
\usepackage{subfig}
\begin{document}

\title{Task-agnostic Decision Transformer for Multi-type Agent Control with Federated Split Training\\
\thanks{$\dag$ Equal contribution}
\thanks{$\ddag$ Work done as an intern at Ping An Technology (Shenzhen) Co., Ltd }
\thanks{$*$ Corresponding author: Xiaoyang Qu (quxiaoy@gmail.com)}
}



\author{
\IEEEauthorblockN{
Zhiyuan Wang\textsuperscript{1$\dag$$\ddag$},
Bokui Chen\textsuperscript{1$\dag$},
Xiaoyang Qu\textsuperscript{2$*$},
Zhenhou Hong\textsuperscript{2},
Jing Xiao\textsuperscript{2},
Jianzong Wang\textsuperscript{2}
}
\IEEEauthorblockA{
\textsuperscript{1}Tsinghua Shenzhen International Graduate School, Tsinghua University, Shenzhen, China\\
Email: wang-zy22@mails.tsinghua.edu.cn, chenbk@tsinghua.edu.cn
}
\IEEEauthorblockA{
\textsuperscript{2}Ping An Technology (Shenzhen) Co., Ltd., Shenzhen, China\\
Email: quxiaoy@gmail.com, {hongzhenhou168, xiaojing661}@pingan.com.cn, jzwang@188.com
}
}

\maketitle

\begin{abstract}
With the rapid advancements in artificial intelligence, the development of knowledgeable and personalized agents has become increasingly prevalent. However, the inherent variability in state variables and action spaces among personalized agents poses significant aggregation challenges for traditional federated learning algorithms. To tackle these challenges, we introduce the Federated Split Decision Transformer (FSDT), an innovative framework designed explicitly for AI agent decision tasks. The FSDT framework excels at navigating the intricacies of personalized agents by harnessing distributed data for training while preserving data privacy. It employs a two-stage training process, with local embedding and prediction models on client agents and a global transformer decoder model on the server. Our comprehensive evaluation using the benchmark D4RL dataset highlights the superior performance of our algorithm in federated split learning for personalized agents, coupled with significant reductions in communication and computational overhead compared to traditional centralized training approaches. The FSDT framework demonstrates strong potential for enabling efficient and privacy-preserving collaborative learning in applications such as autonomous driving decision systems. Our findings underscore the efficacy of the FSDT framework in effectively leveraging distributed offline reinforcement learning data to enable powerful multi-type agent decision systems.
\end{abstract}


\section{Introduction}

Artificial intelligence (AI) has undergone a period of rapid development and growth in recent times, with notable advancements in decision-making and planning. In the domain of self-driving vehicles, AI plays a crucial role in enabling autonomous navigation, obstacle avoidance, and decision-making. However, the coordination and management of multiple intelligent vehicles in complex traffic scenarios pose significant challenges, requiring advanced models that can handle the variability in state variables and action spaces across different agents. Each agent may operate with its own unique set of state variables and action spaces, adding to the complexity of the problem. This necessitates the creation of models that can accommodate such variability. Transformer architecture models \cite{vaswani2017attention} have emerged as efficient and robust solutions \cite{wen2023large}. However, their implementation presents obstacles related to the secure handling of sensitive information and computational efficiency. Offline reinforcement learning models often require centralized training on a single device or server, which can potentially expose sensitive trajectory data distributed across multiple client nodes \cite{pan2019you}. Moreover, the considerable computational demands of model training and data processing make this centralized approach impractical for resource-limited client devices.

The development of intelligent driving systems often involves the collection and processing of sensitive data, such as location information, driving patterns, and user preferences. Ensuring the secure handling of this data while enabling efficient learning for autonomous vehicles is a critical consideration in the field of intelligent driving. Previous studies \cite{such2014survey,hebert2023fedformer,lei2023new} have investigated the protection of sensitive data in reinforcement learning agent systems, emphasizing key challenges that require further attention.

\begin{figure*}[htb]
    \centering
    \subfloat[Single-type agent learning scheme. Traditional centralized training has limitations like privacy concerns and resource bottlenecks as all data is processed in one system. In contrast, split training enhances privacy by keeping sensitive data local. It allows specialized local model training tailored to each agent. Only critical updates are shared globally, reducing communication overhead.\label{fig:single_agent_traditional}]{
        \includegraphics[width=0.95\textwidth]{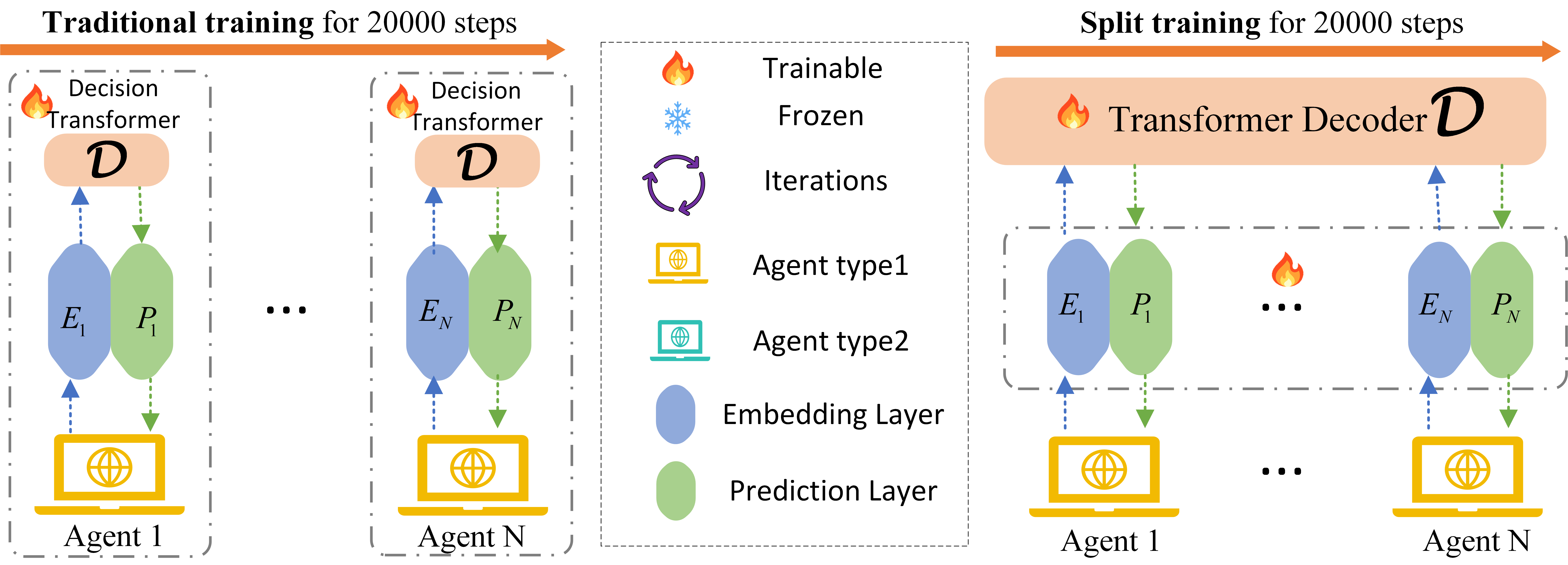}
    }

    \subfloat[A multi-type agent learning scheme with heterogeneous agents. Each agent operates a local model for personalized data. Simultaneously, the server consolidates the updates to generate a unified global model. The server sends global parameters to initialize local models, and clients return updates to synchronize the global model, enabling effective multi-agent control while preserving agent specificity and privacy.\label{fig:multi_agent_split}]{
        \includegraphics[width=0.95\textwidth,height=0.3\linewidth]{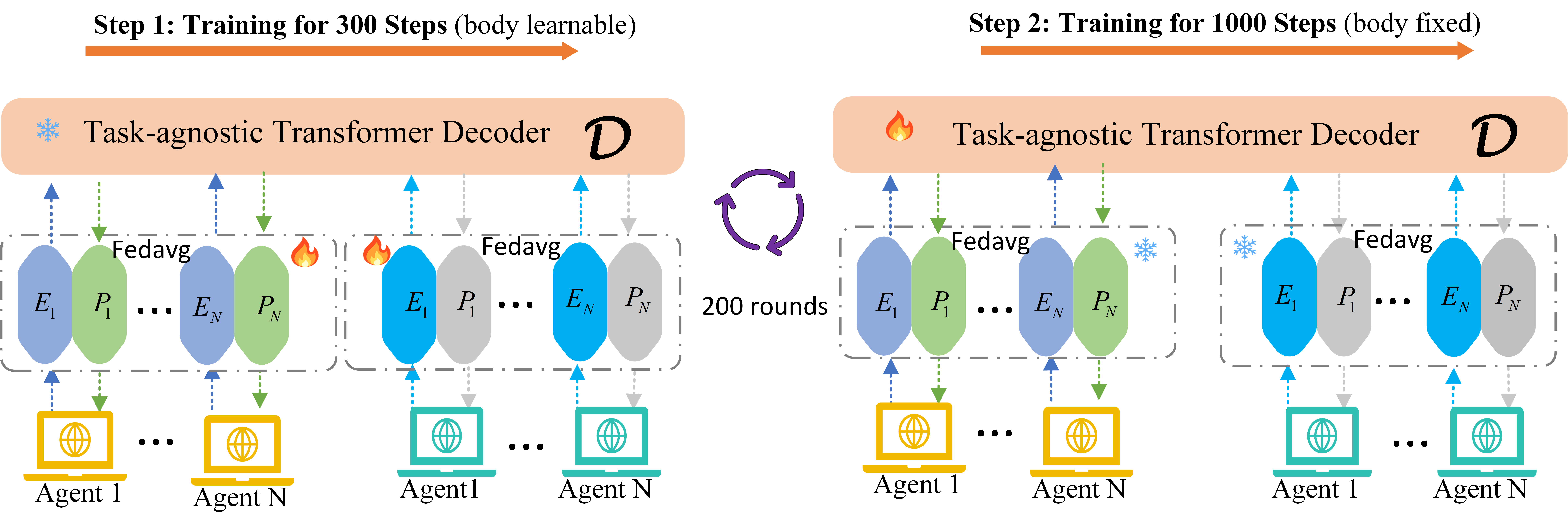}
    }

    \caption{Single-type agent learning task vs. Multi-type agent learning task. }
    \label{fig:combined}
\end{figure*}


Our model introduces a server-side transformer decoder, designed to enhance the efficiency of the learning process.  In offline reinforcement learning, data is often dispersed across clients, making collection challenging. We employ split federated learning (SFL) algorithms to leverage this distributed data for training without central aggregation. This setup places the computationally intense Transformer component on the server, while client-side nodes execute the less demanding but crucial embedding operations. This architecture enhances efficiency by processing only aggregated data from clients while harnessing the capabilities of Transformers for advanced data analysis and decision-making. It also has the potential to enhance the performance and scalability of intelligent driving systems. By integrating split learning into the Decision Transformer architecture, we can effectively utilize distributed offline reinforcement learning datasets.

Our study makes the following contributions:
\begin{itemize}
\item We present FSDT, an innovative framework for federated split learning in continuous control tasks, employing a two-stage training process with a Transformer architecture. 
\item We incorporate a server-side Transformer decoder. This decoder, agnostic to the agent type, processes inputs from various agents without needing their specific details.
\item We limit context length to curtail computational and communication costs of FSDT further. 
\end{itemize}

\section{Related Work}

\textbf{Decision-making tasks with transformer.} Previous work \cite{chen2021decision,lee2022multi,zheng2022online,xu2022hyper,wang2024p2dt} has employed Decision Transformers for decision-making tasks. These approaches facilitate the management of both continuous and discrete action spaces. However, training these Transformers demands large volumes of task-specific offline data via reinforcement learning algorithms. Although this data may be suitable for specific environments and intelligent agents, it often fails to generalize across a broad spectrum of intelligent agents. The data transmission also poses significant privacy breach risks, undermining privacy protection efforts. 

\textbf{Federated Learning Methods.} In response to increasingly stringent data privacy regulations, federated learning has emerged as a prominent approach due to its intrinsic property of not transferring raw data. Nevertheless, there is a shortage of federated learning algorithms explicitly tailored for Transformer architectures. Conventional federated learning algorithms \cite{mcmahan2017communication,li2020federated,bonawitz2019towards,fallah2020personalized,smith2017federated,liu2023fedet,qu2020quantization} generally necessitate uniform model structures across aggregation nodes. Consequently, the globally trained models derived from these algorithms often struggle with personalized tasks in specific environments. Recently, many works \cite{fallah2020personalized,shysheya2022fit} have started focusing on task-agnostic federated learning, with the goal of developing a versatile model capable of accommodating various downstream applications. This constraint restricts the applicability of federated learning, especially in terms of accommodating intelligent agents within specialized environments.

\textbf{Split Learning Methods.} 
Unlike FL, which necessitates that clients fully train models on their local devices and only exchange updates to the model, split learning permits partitioning the training process between client and server \cite{vepakomma2018split,wu2023split}. Such partitioning of responsibilities means clients only need to compute a portion of the model. This lowers their computational workload and memory usage - a vital benefit for devices with constrained resources. Additionally, Split Learning enhances privacy protection as it only involves transmitting intermediate representations or gradients to the server for subsequent processing, excluding the direct transmission of raw data. Recent studies \cite{thapa2022splitfed,abedi2023fedsl,yao2022privacy} have introduced novel methods that combine the benefits of Federated Learning and Split Learning, enabling parallel processing across distributed clients while maintaining model privacy through network splitting and patch shuffling techniques. These approaches aim to achieve efficient training on decentralized sequential data using various architectures such as RNNs and Transformers. However, these approaches have some limitations when applied to multi-type agent scenarios. They do not explicitly address the heterogeneity in state and action spaces across different agent types.



\section{The Proposed Method}
\label{sec:system}

\subsection{Problem Formulation}
\label{ssec:prob}

This study addresses the challenge of training multiple intelligent agents, which could be from different categories, under a federated learning framework. Our Federated Split Decision Transformer (FSDT) framework aims to process and learn from the decentralized and heterogeneous offline data that these agents generate. We define our system environment as comprising a set of $N$ intelligent agents, where each agent is an instance of one of $K$ distinct agent types. These types are denoted as $\{k_n\}_{n=1}^N$, with each type characterized by its unique state space $S_k$ and action space $A_k$. Offline data, which includes action-state trajectories, is utilized from reinforcement learning algorithms to train these agents.

\begin{figure}[htb]
      \centering
      \includegraphics[width=1\linewidth,keepaspectratio]{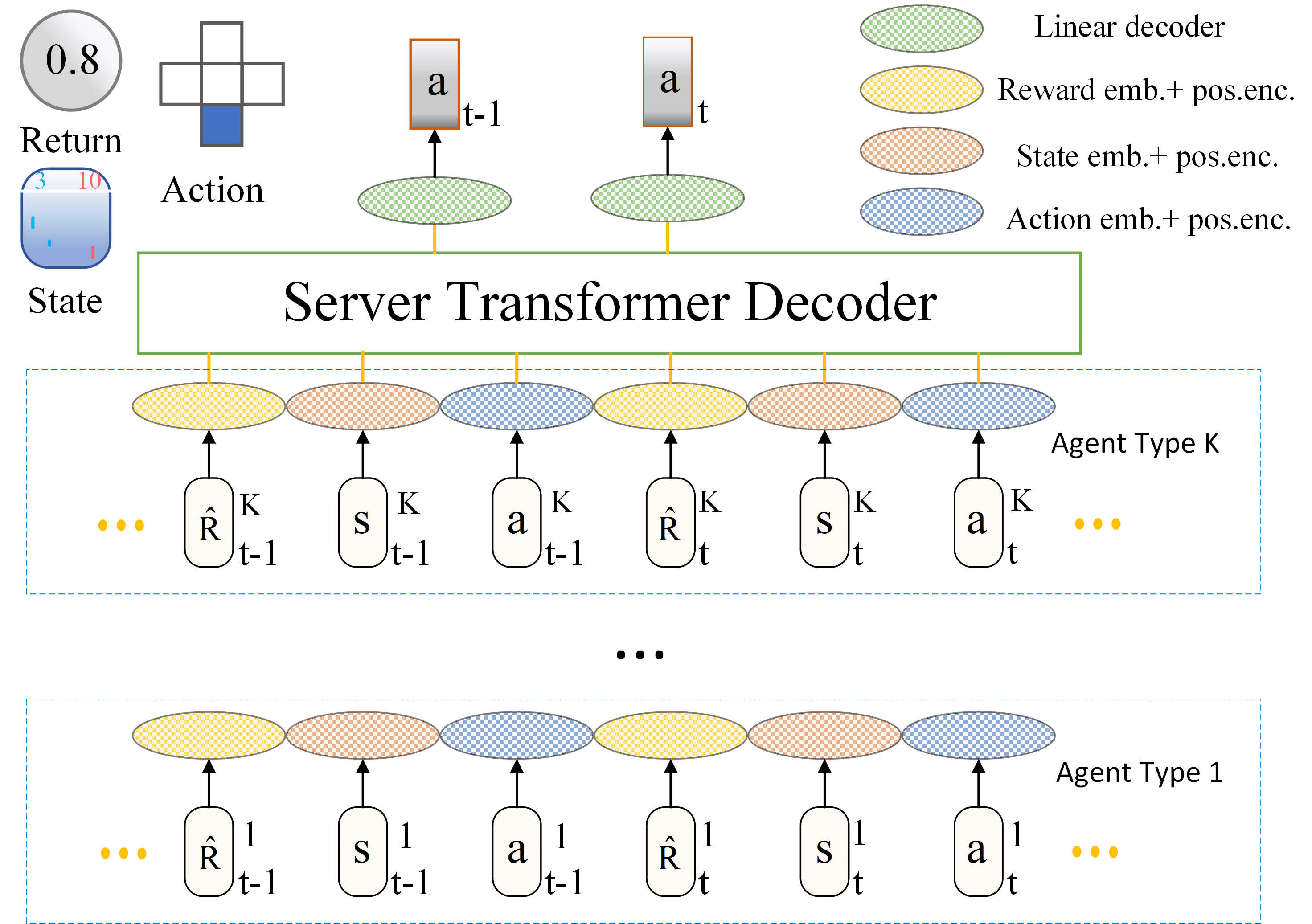}
      \caption{ The core components of the Federated Split Decision Transformer (FSDT). Local agents train with their data, generating embeddings that reflect key features pertinent to decision-making tasks. These embeddings are sent to a server-side transformer decoder, synthesizing the information across all agents to predict actions.}
      \label{fig:env}

\end{figure}

Figure \ref{fig:env} outlines the architecture of FSDT, which significantly differs from traditional centralized training approaches. Each agent $k_n$ independently trains a local model consisting of an embedding module $E^{k_n}$ and a prediction module $P^{k_n}$. These modules incorporate information from trajectories of length $T$, including rewards-to-go, observations, and actions. The output is a 128-dimensional embedding transmitted to a centralized server with a Transformer decoder. This server-side decoder synthesizes the received embeddings from different agent types. It predicts actions by modeling them as Gaussian-distributed vectors, enhancing exploration and learning stability.

The learning mechanism for each agent type $k$ can be conceptualized as contextual learning within a Markov Decision Process (MDP), formalized by $(S^k, A^k, P^k, R^k)$. Here, $S^k$ represents the state space for agent type $k$, with each state $s \in S^k$, while $A^k$ is the corresponding action space with actions $a \in A^k$. The transition dynamics are captured by $P^k(s'|s, a)$, and the reward function is represented as $r = R^k(s, a)$. For an agent of type $k$, the state, action, and reward at timestep $t$ are denoted by $s_t^k$, $a_t^k$, and $r_t^k = R_k(s_t^k, a_t^k)$, respectively. The primary objective in reinforcement learning is to find an optimal policy that yields the highest anticipated cumulative rewards, which for agent type $k$ is defined as $E[\sum_{t=1}^{T} r_t^k]$. The returns-to-go are modeled as $\hat{R}_t^k = \sum_{t'=t}^{T} r_{t'}^k$, leading to the trajectory representation conducive to autoregressive training:
\begin{equation}
\tau_k = (\hat{R}_1^k, s_1^k, a_1^k, \hat{R}_2^k, s_2^k, a_2^k, . . . , \hat{R}_T^k, s_T^k, a_T^k)
\end{equation}
This formulation allows us to address the heterogeneity in agent types and their data while preserving privacy and leveraging the shared learning capabilities of federated learning.

\subsection{Algorithm Architecture }
\label{ssec:arch}

\begin{figure}[htb]
    \centering
    \includegraphics[width=1\linewidth,keepaspectratio]{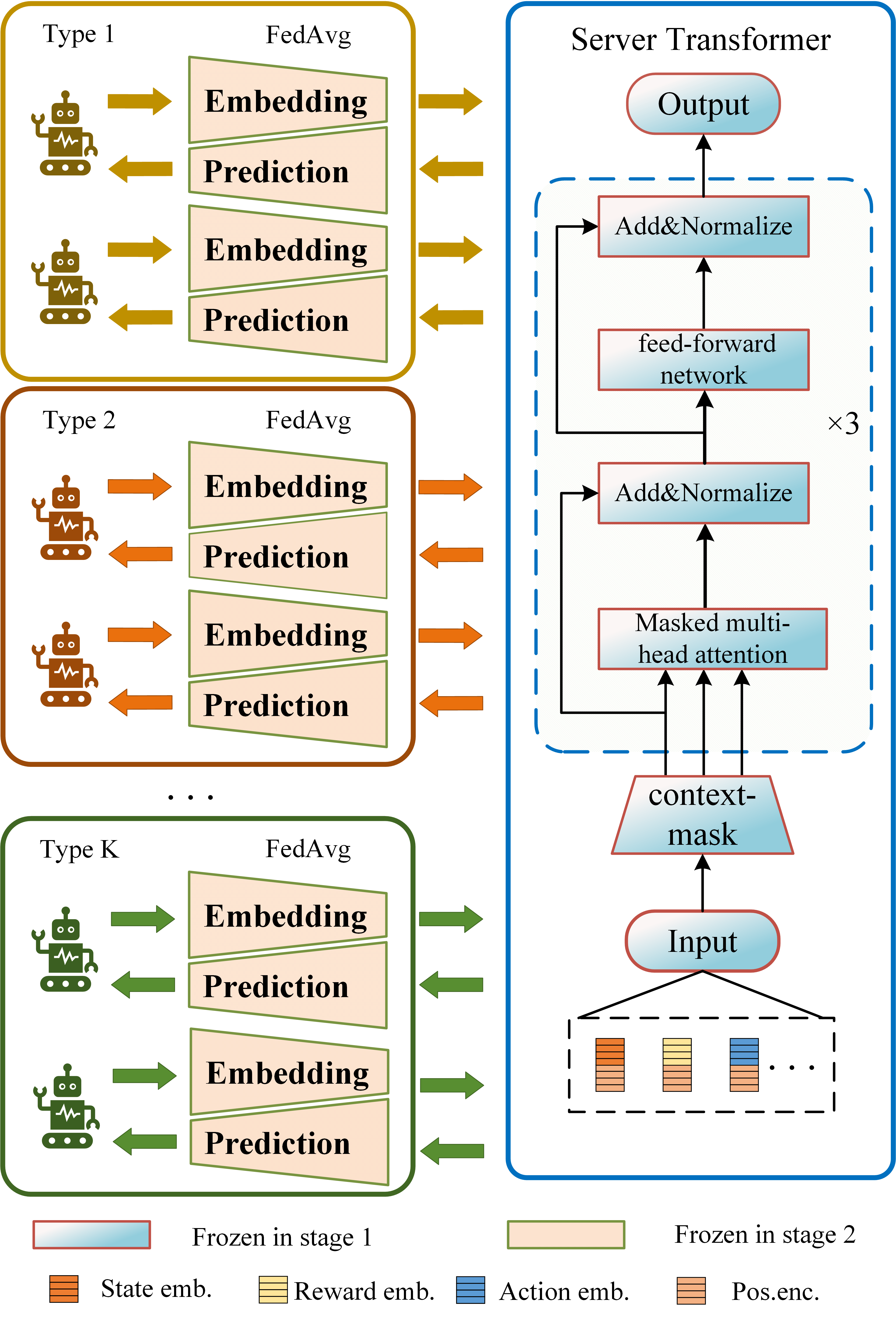}

    \caption{The framework of our algorithm. At the personalized client agent,  we train the embedding and prediction models in stage 1. On the server side, we introduce the Transformer decoder without the embedding layer in stage 2.}
    \label{fig:framework}

\end{figure}

Fig~\ref{fig:framework} illustrates our novel approach of integrating the Decision Transformer architecture into a federated learning framework, which stands in contrast to traditional centralized techniques. This integration tackles the unique challenges posed by distributed data sources, enabling a learning process that is both more effective and adaptable.

The temporal steps and reward values are modeled as a sequence encapsulating temporal and spatial information. This sequence,  limited to a length of $h$, is fed into the local models of the intelligent agents. The input is transformed through a personalized embedding model, with the dimensionality of the input vector defined as $Q=d\times h+b\times h+h$. Within this framework, the variable $d$ denotes the dimensions of the state, while $b$ signifies the dimensions of the action. The output from this embedding model is a 128-dimension hidden vector.

At the agent level, personalized models $E_{t}^{k_n}$ and $P_{t}^{k_n}$ are in operation. The embedding model $E_{t}^{k_n}$ take the past $m$ timesteps of reward-to-go $q_{t-m:t}$, observation $s_{t-m:t}$, and action $a_{t-m:t}$. The input tokens are produced by mapping the inputs to a 128-dimensional embedding space, followed by the addition of a timestep embedding $\omega(t)$. The embedding output $u_{r_t}^{k_n}$, $u_{s_t}^{k_n}$ and $u_{a_t}^{k_n}$ can be defined as:

\begin{align}
u_{q_t}^{k_n} &= \phi_r(r_t^{k_n}) + \omega(t) \\
u_{s_t}^{k_n} &= \phi_s(s_t^{k_n}) + \omega(t) \\
u_{a_t}^{k_n} &= \phi_a(a_t^{k_n}) + \omega(t) 
\end{align}

The central server hosts a Transformer decoder (${{G}^{t}}$), with the embedding layer removed, and this model receives inputs from the embedding module. It then predicts the next tokens similarly to the Generative Pre-Training model \cite{radford2019language}. The output token, denoted as $v_{q_t}^{k_n}$, $v_{s_t}^{k_n}$ and $v_{a_t}^{k_n}$, is defined as:

\begin{equation}
 v_{q_t}^{k_n}, v_{s_t}^{k_n}, v_{a_t}^{k_n} = G(u_{q_{t-m}}^{k_n} , u^{k_n}_{s_{t-m}} , u_{a_{t-m}}^{k_n} , \ldots , u_{q_t}^{k_n}, u_{s_t}^{k_n}, u_{a_t}^{k_n})
\end{equation}

The prediction model $P_{t}^{n_k}$ then converts the output from the server into action vectors. Similar to SAC \cite{haarnoja2018soft}, instead of predicting deterministic actions, we predict a Gaussian distribution over actions based on the output tokens from the server, for improved exploration and more stable learning.

\begin{equation}
\pi_\theta (a_t^{k_n} | v_{s_t}^{k_n}) = \mathcal{N} (\mu_\theta (v_{s_t}^{k_n}) , \Sigma_\theta (v_{s_t}^{k_n})) \\
\end{equation}
The covariance matrix $\Sigma_\theta$ is assumed to be diagonal in the equation above. The goal of training is to reduce the negative logarithm of the likelihood that the model will produce the correct action.

\subsection{Training Procedure}
\label{ssec:procedure}

Inspired by \cite{qu2022rethinking,park2022multi,park2021federated} in Computer Vision, our training procedure (Algorithm \ref{alg:federated_learning}) encompasses two stages. Initially, the transformer parameters ${{g}_{t}} $ on the server side are kept constant. The parameters of the global embedding and prediction models from the previous iteration, $e_{k}^{t-1}$ and $p_{k}^{t-1}$, are then distributed to the users. The distribution of these parameters is conducted based on the class of the intelligent agent. After that, the parameters of the embedding and prediction networks denoted as  $e^{k_n}_{t}$ and $p^{k_n}_{t}$ respectively, are optimized as follows:
\begin{equation}
\underset{e^{k_n}_{t},p^{k_n}_{t}}{\mathop{\min }}\,\sum\limits_{k=1}^{K}{\sum\limits_{n=1}^{{{N}_{k}}}{{{l}_{k}}(y^{k_n}_{t}}},P^{k_n}_{t}({{G}^{t}}(E^{k_n}_{t}(x^{k_n}_{t}))))
\end{equation}

\begin{algorithm}[htb]
\caption{FSDT: FEDERATED SPLIT DECISION TRANSFORMER}
\label{alg:federated_learning}
\begin{algorithmic}[1]
\State \textbf{Input:} Initial global models $g_0^k$ for all $k$ in the range $(0, K]$, initial server model $v_0$, initial personal local models $w_{0,n}^k$
\State \textbf{Output:} Final global model $G_C^k$, final server model $v_C$

\For{$c=1$ \textbf{to} $C$}
\For{$k=1$ \textbf{to} $K$}
\For{$n=1$ \textbf{to} $N_k$}
\State $w_{t}^{k_n} \gets g_{t}^{k_n}$
\State $w_{t}^{k_n} \gets \text{LocalUpdate}(w_{t}^{k_n}, v_{t-1})$
\EndFor
\State $G_t^k \gets \text{GlobalUpdate}(w_{t}^{k_1}, w_{t}^{k_2}, \dots, w_{t}^{k_N}, v_{t-1})$
\EndFor
\State $v_t \gets v_{t-1}$
\For{$k=1$ \textbf{to} $K$}
\For{$n=1$ \textbf{to} $N$}
\State $v_t \gets \text{TrainServer}(G_t^k, v_t)$
\EndFor
\EndFor
\EndFor
\end{algorithmic}
\end{algorithm}

Furthermore, model aggregation is executed among nodes of identical intelligent agents using Equation (1), resulting in the derivation of global models $E_{t}^{k}$ and $P_{t}^{k}$. The parameters of these models are calculated as follows:
\begin{equation}
e_{t}^{k}=\frac{1}{{{N}_{k}}}\sum\limits_{n=1}^{{{N}_{k}}}{e^{k_n}_{t}}
\end{equation}

\begin{equation}
p^{k}_{t}=\frac{1}{{{N}_{k}}}\sum\limits_{n=1}^{{{N}_{k}}}{p^{k_n}_{t}}
\end{equation}

In the second stage, all client-side model parameters $E^{k_n}_{t}$ and $P^{k_n}_{t}$ are kept frozen, while the server-side model ${{G}_{t}}$ undergoes training utilizing the corresponding dataset. This process remains agnostic to the types of intelligent agents involved. The optimization objective of this training process is given by:
\begin{equation}
\underset{{{g}_{t}}}{\mathop{\min }}\,\sum\limits_{k=1}^{K}{\sum\limits_{n=1}^{{{N}_{k}}}{{{l}_{k}}(y^{k_n}_{t}}},P^{k_n}_{t}({{G}^{t}}(E^{k_n}_{t}(x^{k_n}_{t}))))
\end{equation}

The introduced two-phase training process and the use of a Transformer decoder on the server side are conceptually designed to enhance the system's capabilities. The Transformer decoder is designed to integrate information from diverse agents more effectively, optimizing the learning process. To verify this, we conducted a small-scale experiment showing performance gains in simulated tasks by utilizing the Transformer decoder compared with conventional methods.

This split process enables balanced and efficient model training, allowing global model updates without incurring excessive communication costs. The implementation of local updates incorporates the unique characteristics of each agent. Conversely, the global updates enable the entire learning process to benefit from the collective knowledge that is common among all agents. This method successfully reconciles the need for personalized learning and the gains of collective intelligence.

\begin{figure*}[htb]
    \centering
    \subfloat[The D4RL score of 3 kinds of agent with global model on Medium-Expert dataset.\label{fig:sub1}]{
        \includegraphics[width=.3\linewidth]{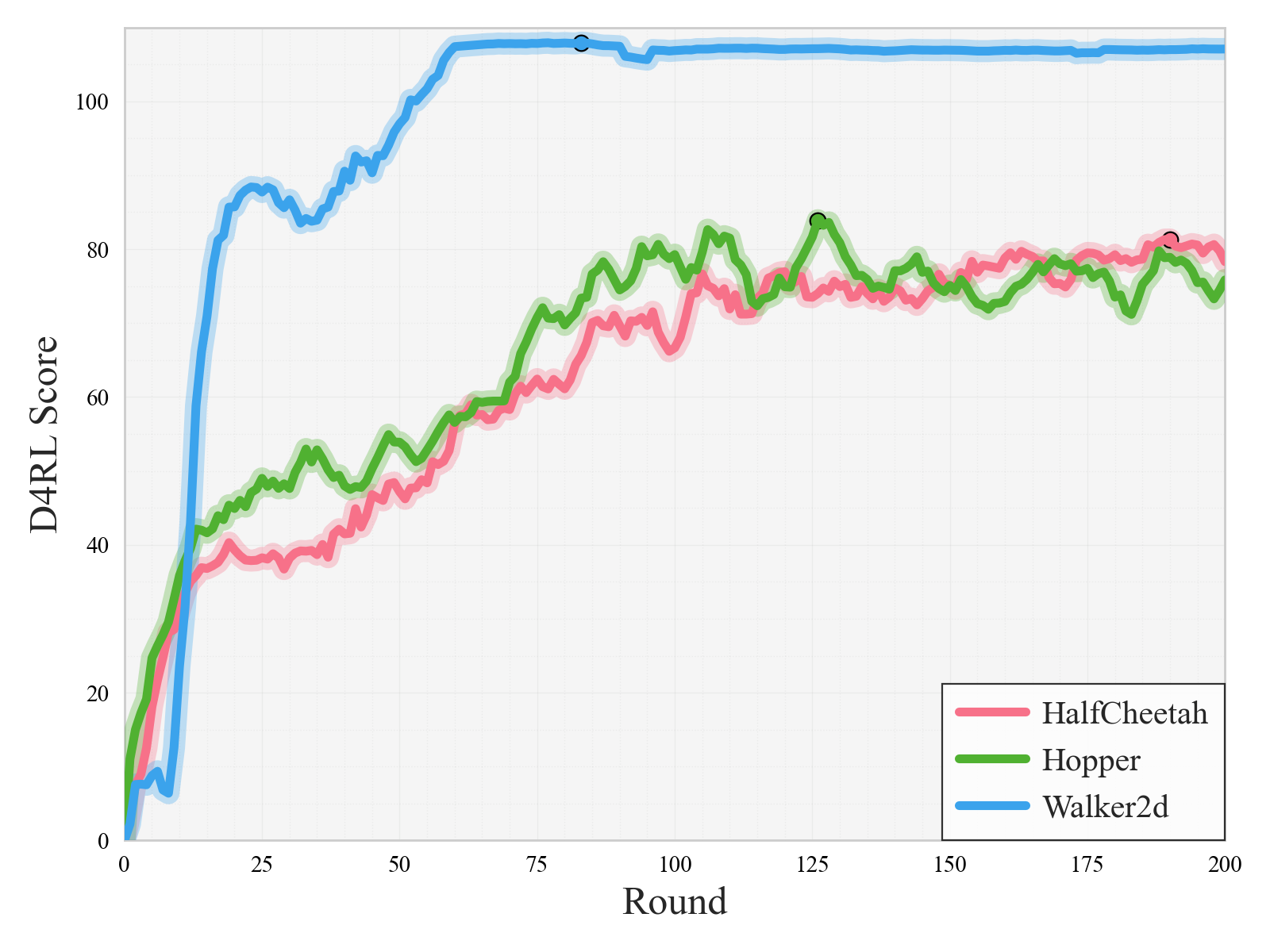}
    }
    \hfill
    \subfloat[The D4RL score of 3 kinds of agent with global model on Medium dataset.\label{fig:sub2}]{
        \includegraphics[width=.3\linewidth]{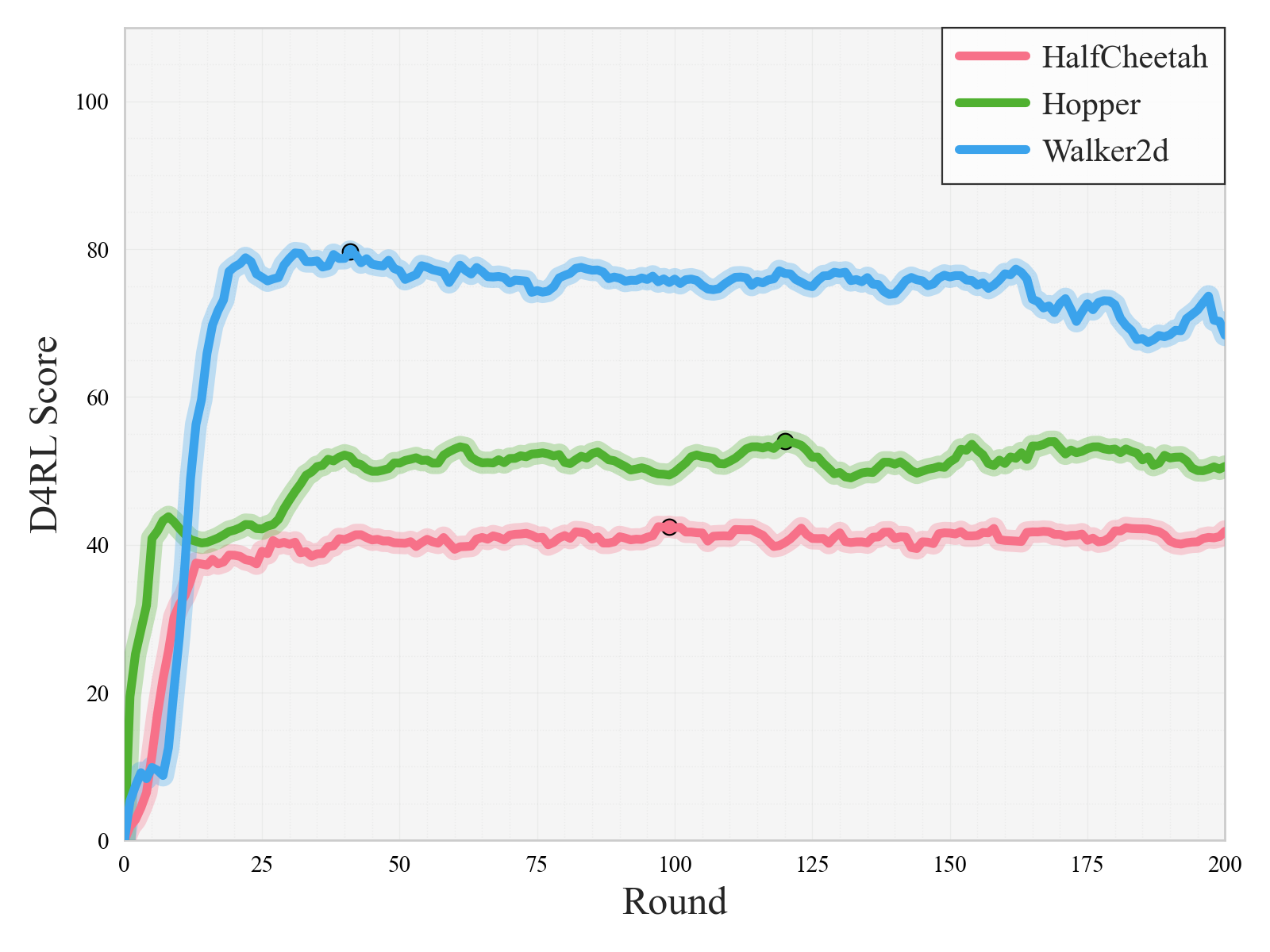}
    }
    \hfill
    \subfloat[The D4RL score of 3 kinds of agent with global model on Medium-Replay dataset.\label{fig:sub3}]{
        \includegraphics[width=.3\linewidth]{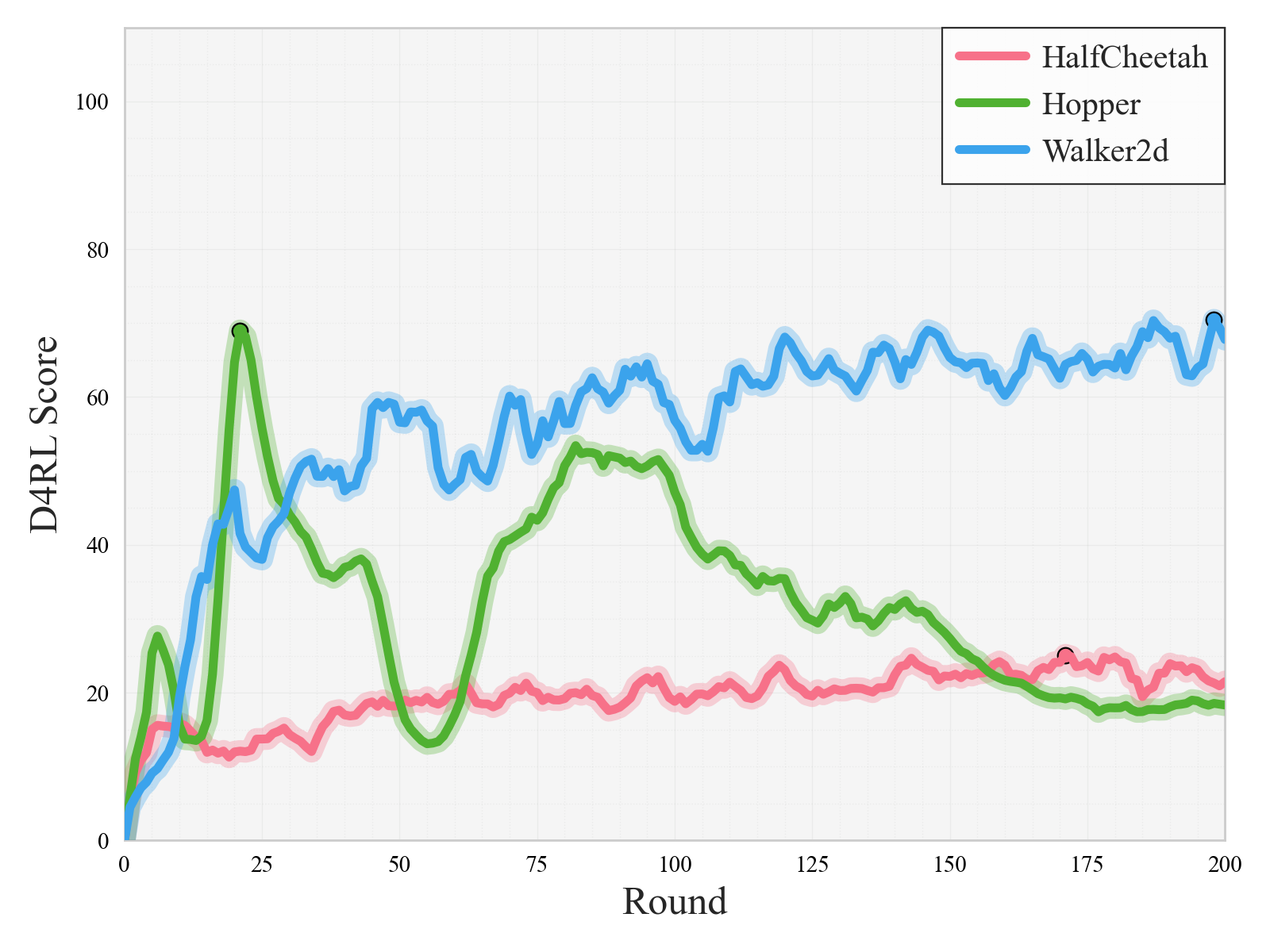}
    }
    \caption{Performance comparison of different agents.}
    \label{fig:performance_comparison}
\end{figure*}

By adopting this approach, our research leverages the strengths of distributed computing to address the challenges posed by continuous control tasks in federated learning environments. The server, equipped with the Transformer's powerful data processing capabilities, handles the heavy computational load. In contrast, the client nodes are each adapted to their unique task requirements. They manage the embedding processes which require less computational power but are critical for addressing the specific needs of the tasks.

\section{evaluation}
\label{sec:eval}
\subsection{Experiment Setup }
\label{ssec:exp}

We assessed the FSDT algorithm using the Mujoco simulator with three robot control environments: HalfCheetah, Hopper, and Walker2D, utilizing the D4RL dataset. These environments were chosen as they represent a diverse set of continuous control
tasks with varying complexity, allowing for a comprehensive evaluation of our approach. The experiment involved 30 agents: 10 HalfCheetah, 10 Hopper, and 10 Walker2D. The D4RL dataset, which includes expert, medium, and medium replay levels, was partitioned among the agents in accordance with federated learning principles. This ensured the data allocation was independent and identically distributed (IID.), with each agent receiving a randomly selected subset from the overall distribution in proportion to the total examples. Notably, the distribution of the three levels of data (Medium-Expert, Medium, and Medium-Replay) was approximately equal among agents of the same type.

In the training process, there were 200 rounds of communication between the clients and server. In each round, every intelligent agent type first underwent 300 steps of local training on the client side. After that, there were 1000 steps of training on the server side to consolidate the learning across all agents in the federated network.


\subsection{Results Analysis}
\label{ssec:analysis}

This study employed the D4RL score as the evaluation metric, serving as a yardstick for comparing and contrasting results. Table \ref{tab:experiment_results} presents a comparison of our proposed algorithm against multiple established techniques, such as DT\cite{chen2021decision}, CQL\cite{kumar2020conservative}, BEAR-v\cite{kumar2019stabilizing}, BRAC\cite{wu2019behavior}, AWR\cite{peng2019advantage}, and behaviour cloning (BC).

\begin{table*}[htbp]
\centering
\caption{Results for D4RL datasets. The D4RL score of our proposed algorithm FSDT and D4RL score using other six different methods: DT, CQL, BEAR, BRAC-v, AWR, and BC.}
\resizebox{\textwidth}{!}{
\footnotesize
\begin{tabular}{@{}lcccccccc@{}}
\toprule
\textbf{Dataset} & \textbf{Environment} & \textbf{DT} & \textbf{CQL} & \textbf{BEAR} & \textbf{BRAC-v} & \textbf{AWR} & \textbf{BC} & \textbf{Ours} \\
\midrule
Medium-Expert  & HalfCheetah & 86.8  & 62.4 & 53.4 & 41.9 & 52.7 & 59.9 & 84.5\\
Medium-Expert  & Hopper      & 107.6  & 111.0 & 96.3 & 0.8  & 27.1 & 79.6 & 89.1\\
Medium-Expert  & Walker      & 108.1  & 98.7 & 40.1 & 81.6 & 53.8 & 63.6 & 108.1\\
Medium-Expert  & Average     & 100.8  & 90.7 & 63.3 & 41.4 & 44.5 & 67.7 & 93.9 \\
\midrule
Medium        & HalfCheetah & 42.6  & 44.4 & 41.7 & 46.3 & 37.4 & 43.1 & 43.3\\
Medium        & Hopper      & 67.6   & 58.0 & 52.1 & 31.1 & 35.9 & 63.9 & 57.5\\
Medium        & Walker      & 74.0  & 79.2 & 59.1 & 81.1 & 17.4 & 77.3 & 81.0\\
Medium        & Average     & 61.4 & 60.5 & 51.0 & 52.8 & 30.2 & 61.4 & 60.6 \\
\midrule
Medium-Replay & HalfCheetah & 36.6 & 46.2 & 38.6 & 47.7 & 40.3 & 4.3 & 28.9\\
Medium-Replay & Hopper      & 82.7& 48.6 & 33.7 & 0.6 & 28.4 & 27.6 & 73.6\\
Medium-Replay & Walker      & 66.6 & 26.7 & 19.2 & 0.9 & 15.5 & 36.9 & 76.3\\
Medium-Replay & Average     & 62.0 & 40.5 & 30.5 & 16.4 & 28.1 & 22.9 & 59.6 \\
\midrule
\textbf{Average(All Settings)} & & 74.7 & 63.9 & 48.2 & 36.9 & 34.3 & 46.4 & 71.4 \\
\bottomrule
\label{tab:experiment_results}
\end{tabular}}
\end{table*}

We computed the mean performance of the listed algorithms across the expert, medium, and medium-replay datasets. The results, summarized in Table \ref{tab:experiment_results}, demonstrate that our FSDT algorithm under federated split learning settings surpasses the majority of other methods and achieves performance comparable to DT in non-federated scenarios.

\begin{table}[htb]
\caption{Parameter Analysis of FSDT vs DT at the client}
\label{tab:parameter size}
\centering
\normalsize
\begin{tabular}{cccccc}
\toprule
\textbf{Method} & \textbf{Agent} & \textbf{Part} & \textbf{Param.} & \textbf{Size (MB)} \\
\midrule
\multirow{3}{*}{DT} & HalfCheetah & Total & 27.73M & 28.53 \\
& Walker2D & Total & 27.73M & 28.53  \\
& Hopper & Total & 27.72M & 28.52 \\
\midrule
\multirow{6}{*}{Ours} & \multirow{2}{*}{HalfCheetah} & Emb. & 131.7k & 0.502  \\
& & Pred. & 3.1k & 0.012  \\

& \multirow{2}{*}{Walker2D} & Emb. & 131.7k & 0.502  \\
& & Pred. & 3.1k & 0.012  \\

& \multirow{2}{*}{Hopper} & Emb. & 130.6k & 0.498  \\
& & Pred. & 1.9k & 0.007  \\
\bottomrule
\end{tabular}
\end{table}



\subsection{Experiment Analysis}
\label{ssec:consumption analysis}

We performed a consumption analysis on the FSDT model, with a specific emphasis on the number of parameters, as presented in Table \ref{tab:parameter size}. The FSDT employs a context-truncated transformer decoder model, leading to a reduced parameter count compared to the decision transformer strategy.



Figures \ref{fig:sub1} to \ref{fig:sub3} show our proposed algorithm FSDT performance trends as communication rounds increases. It can be observed that around 100 rounds of training, the model converges basically. After that, if continuing training, due to overfitting issues, the model's performance on some datasets may decrease slightly.

\begin{figure}[htb]
    \centering
    \subfloat[The D4RL score of three agents via different clients number on Medium-Expert dataset.\label{fig:sub4}]{
        \includegraphics[width=.45\linewidth]{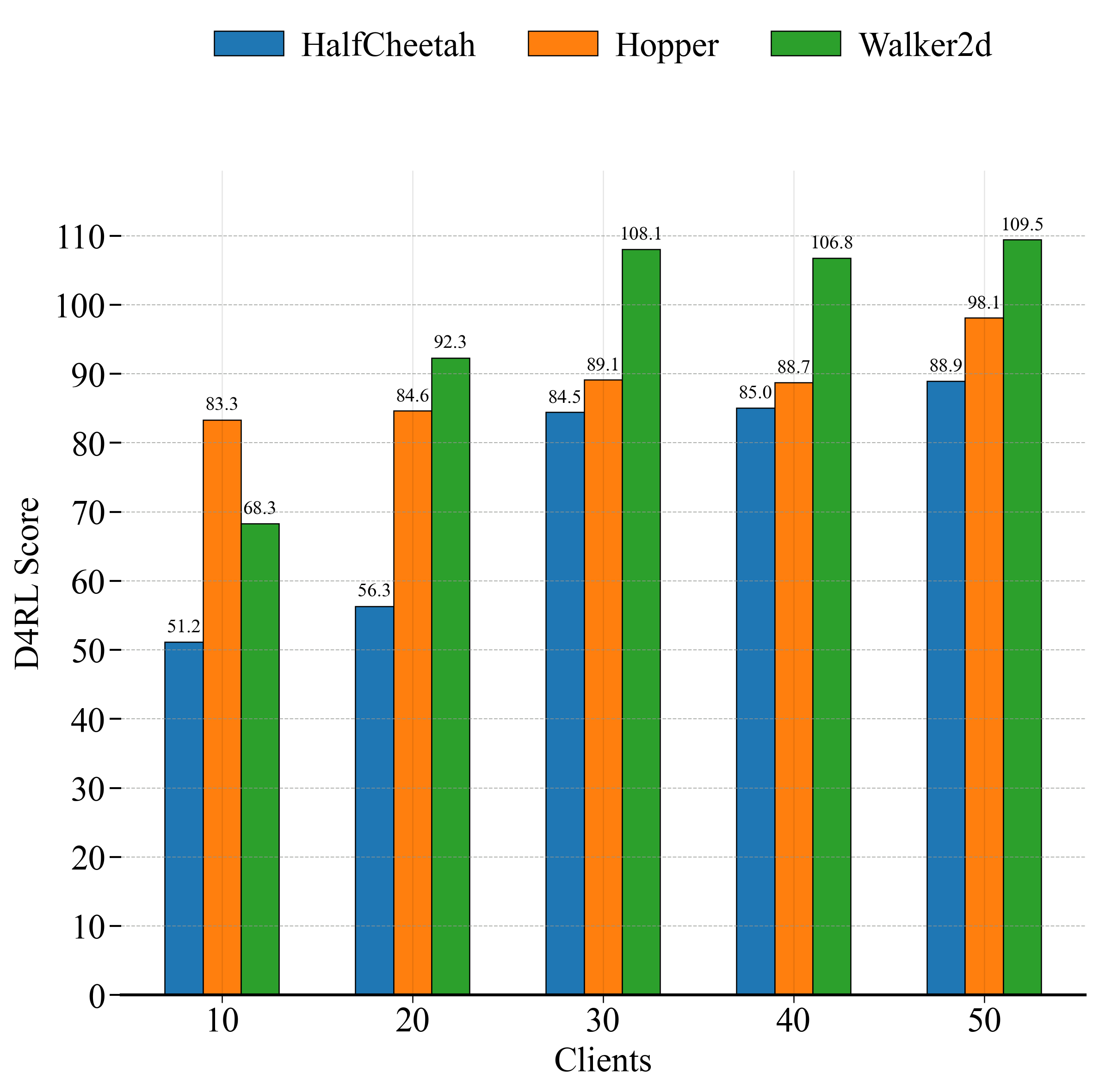}
    }
    \hfill
    \subfloat[Average D4RL score and standard variation via different context length on Medium-Expert dataset.\label{fig:sub5}]{
        \includegraphics[width=.45\linewidth]{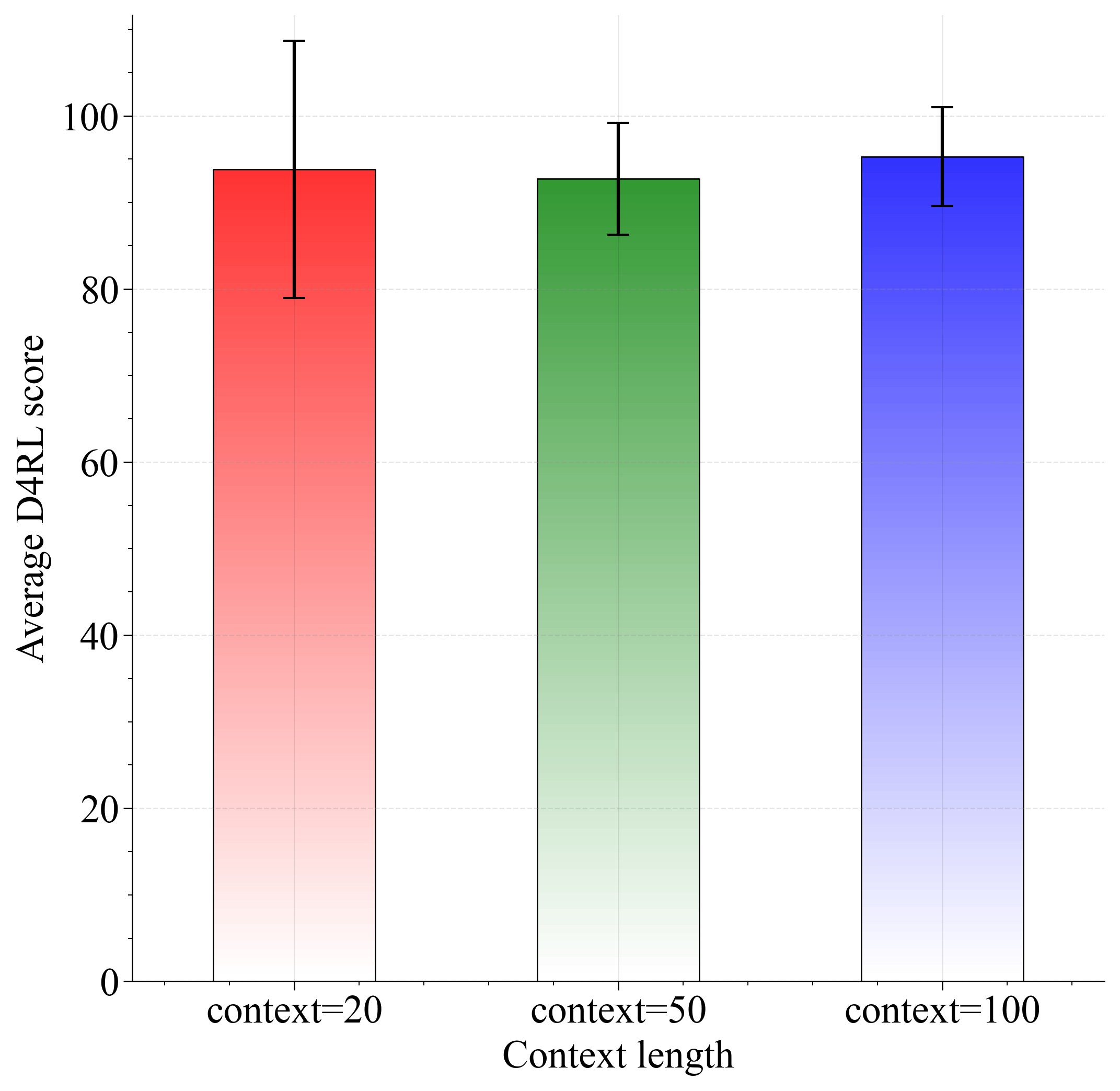}
    }
    \caption{The ablation experiment results of our proposed algorithm FSDT on Medium-Expert dataset.}
    \label{fig:res}
\end{figure}

In Figure \ref{fig:sub4}, when clients' number is below 30, the model is relatively insufficiently trained, leading to lower accuracy. For each client number, the distributions of the three agent types are almost equal. As illustrated in Figure \ref{fig:sub5}, imposing a limit on the context length does not considerably influence the model's final performance. However, it can lead to a notable enhancement in computational efficiency.

Moreover, a significant portion, approximately 85\%, of the model parameters are allocated on the server side. This configuration can effectively reduce client devices' communication and computational overhead, enabling a more efficient learning process with less resource requirement. 

The operations in the Transformer decoder and the embedding and prediction models primarily determine the computational complexity of FSDT. The amount of data transmitted is primarily influenced by the quantity of participants and the dimensions of the models involved. It is reduced by conducting local updates on the client side and only sharing model parameter updates with the server instead of full model parameters.

Our results primarily showcase the performance enhancements achieved through the novel implementation of a server-side Transformer decoder in a split learning context. The enhanced performance in the results suggests that the model is more efficiently learning from the distributed data. This more efficient data handling may potentially lead to privacy improvements, as less private data needs to be exposed during training to achieve good performance.

\section{Conclusion}
\label{sec:typestyle}

Throughout the research, we introduced a novel split offline reinforcement learning approach, the FSDT, explicitly designed to cater to the complexities of personalized intelligent agents. Our empirical results underscored the effectiveness of this approach, which delivered high performance while minimizing overhead. The computational efficiency of FSDT is of utmost importance, as it enables clients with limited hardware resources to engage in federated learning, a feat that would otherwise pose substantial challenges. This makes it an ideal solution for agents operating under resource constraints. Future
research directions include extending FSDT to handle more complex agent architectures and exploring applications in real-world scenarios such as autonomous driving and robotics.

\section*{Acknowledgment}
This work was supported by the Tsinghua-Toyota Joint Research Fund (Grant No. 20223930089),  and the Tsinghua Shenzhen International Graduate School Fund (HW2020005, JC2021009).



\end{document}